\documentclass[conference]{IEEEtran}
\IEEEoverridecommandlockouts
\usepackage{cite}
\usepackage{amsmath,amssymb,amsfonts}
\usepackage{algorithmic}
\usepackage{graphicx}
\usepackage{textcomp}
\usepackage{xcolor}
\usepackage{booktabs}
\def\BibTeX{{\rm B\kern-.05em{\sc i\kern-.025em b}\kern-.08em
    T\kern-.1667em\lower.7ex\hbox{E}\kern-.125emX}}
\begin{document}

\title{Improving mitosis detection on histopathology images using large vision-language models\\
\thanks{$^*$ Work done during an internship at Microsoft Research; corresponding author dingrw@ucla.edu}
}

\author{\IEEEauthorblockN{Ruiwen Ding$^*$ ~~James Hall$^\dag$~~ Neil Tenenholtz$^\dag$~ ~Kristen Severson$^\dag$}\\
\IEEEauthorblockA{$^*$ University of California, Los Angeles, CA, USA}
\IEEEauthorblockA{$^\dag$ Microsoft Research, Cambridge, MA, USA}
}

\maketitle

\begin{abstract}
In certain types of cancerous tissue, mitotic count has been shown to be associated with tumor proliferation, poor prognosis, and therapeutic resistance. Due to the high inter-rater variability of mitotic counting by pathologists, convolutional neural networks (CNNs) have been employed to reduce the subjectivity of mitosis detection in hematoxylin and eosin (H\&E)-stained whole slide images. However, most existing models have performance that lags behind expert panel review and only incorporate visual information. In this work, we demonstrate that pre-trained large-scale vision-language models that leverage both visual features and natural language improve mitosis detection accuracy. We formulate the mitosis detection task as an image captioning task and a visual question answering (VQA) task by including metadata such as tumor and scanner types as context. The effectiveness of our pipeline is demonstrated via comparison with various baseline models using 9,501 mitotic figures and 11,051 hard negatives (non-mitotic figures that are difficult to characterize) from the publicly available Mitosis Domain Generalization Challenge (MIDOG22) dataset.
\end{abstract}

\begin{IEEEkeywords}
Mitosis detection; digital pathology; large vision-language models
\end{IEEEkeywords}

\section{Introduction}
Mitosis, cell division and duplication, is a part of the cell cycle. Mitotic count has been shown to be associated with tumor proliferation and poor prognosis \cite{elston1991pathological} \cite{moreira2020grading} and plays an important role in treatment recommendations for various cancer diagnoses \cite{sledge2016canine}. However, in clinical practice, mitotic counting on routinely-acquired H\&E tissue slides has high inter-rater variability largely due to the difficulties of both identifying the candidate regions with high mitotic activity and classifying the mitotic figures against non-mitotic figures \cite{wei2019agreement}. Several studies have proposed machine learning models for mitosis detection to reduce subjectivity. For instance Bertram et al. show computer-assisted systems improve the reproducibility and accuracy of mitosis detection on H\&E images \cite{bertram2022computer}. Other studies have considered fully automated pipelines. For example, Ji et al. framed mitosis detection as both classification and object detection problems and built predictive models using ResNet-50 and Faster R-CNN architectures \cite{ji2023considerations}. In the MIDOG 2021 challenge, participants developed various CNN-based approaches to detect mitosis, and the F1 score of ensembled top-five models was 0.77 \cite{aubreville2023mitosis}. Cayir et al. proposed a two-stage CNN-based framework called  MITNET to classify mitosis and obtained an F1 score of 0.69 on the MIDOG dataset and 0.49 on an in-house dataset \cite{ccayir2022mitnet}. 

Although early work relying on deep-learning based methods has shown promise, performance still lags behind a panel of experts. Moreover, prior work has largely relied solely on visual information, whereas pathologists typically leverage both visual and natural language data when learning and reasoning through histopathologic concepts. Recently, pre-trained vision-language models have shown promising results on various downstream tasks. CLIP \cite{radford2021learning}, which was trained on large-scale image-caption pairs via contrastive learning, has been shown to have competitive zero-shot transfer results on various downstream tasks when compared to fully-supervised baselines. Several works have shown the potential of using such contrastive learning-based vision-language models in the biomedical domain \cite{huang2023leveraging}\cite{zhang2023large}\cite{lu2023towards}. BLIP is another state-of-the-art vision-language model that not only uses contrastive learning but also incorporates two additional objectives to further align image-text representations \cite{li2022blip}. These losses enable BLIP to excel in both understanding-based tasks such as VQA or visual reasoning and generation-based tasks such as image captioning.

In this work, we introduce a tile-level mitosis classification pipeline using BLIP by framing the problem as both an image captioning problem and a VQA problem with metadata such as tumor and scanner types incorporated in the question (prompt) as context. We demonstrate that BLIP can improve the mitosis detection accuracy as compared to the CNN-based and vision transformer (ViT)-based approaches that only leverage visual information. The main contributions of our work are as follows:

1. To the best of our knowledge, this is the first work that leverages large vision-language models for mitosis detection.

2. We show that incorporating metadata into the question (prompt) improves the prediction accuracy.

3. We compare our proposed approaches with multiple baselines and show that BLIP outperforms single-modality (vision) models as well as the widely used vision-language model CLIP.

\begin{figure*}[]
\centering
\includegraphics[width=0.8\linewidth]{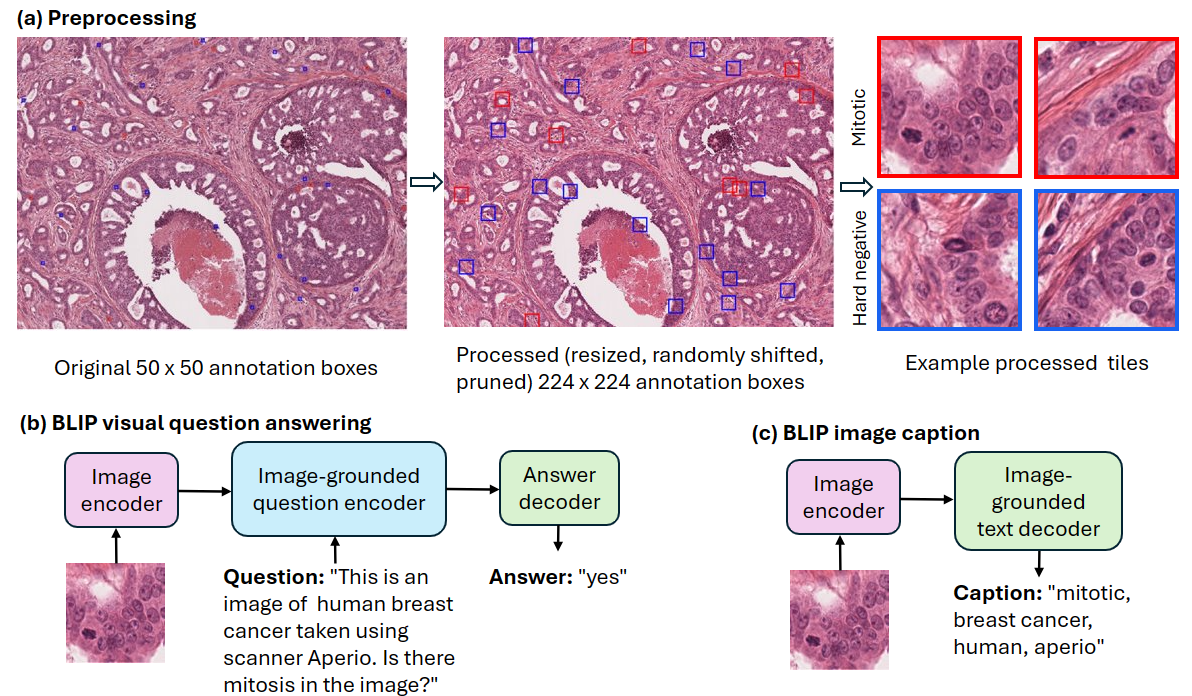}
\caption{The overview of preprocessing and modeling. (a) represents the preprocessing of the MIDOG22 dataset including resizing, random shift, and pruning of the annotation boxes. (b) and (c) represent the model input, output, and three encoder or decoder components of the BLIP VQA model and BLIP image caption model respectively. }
 \label{fig:figure1}
\end{figure*}

\section{Methods}

\subsection{Dataset and preprocessing} \label{sec:dataset}
In this work, we used the MIDOG 2022 challenge dataset (MIDOG22), which includes 354 annotated cases of different tumor types, species, and scanners \cite{aubreville2023mitosis}. The dataset contains annotations for 9,501 mitotic figures and 11,051 hard negatives in the form of 50 by 50 pixel bounding boxes. Bounding box annotations were converted into tiles and tile-level labels as follows. Each box was first expanded to include an area of 224 by 224 pixels and then was randomly shifted in both horizontal and vertical directions from the center up to a fixed threshold. The threshold ($\pm$ 80 pixels) allows the maximum amount of shifting while precluding mitotic (or non-mitotic) figures from being cropped out. Shifts were also constrained to ensure the resulting tile was within the TIFF image boundary. This was performed to introduce variation in the location of the (non-) mitotic figures in the box/tile so that the figures are not always centered in the tile. After the resizing and shifting, any hard negatives that overlap with the positives were discarded to avoid potential false negatives in the shifted boxes. Figure \ref{fig:figure1}a shows before and after the preprocessing of an example case. 

The train (60\%) / validation (20\%) / test (20\%) split was performed at the patient level and was repeated 5 times using different random seeds. As a form of data augmentation, in the training sets, 10 randomly shifted boxes were derived per (non-) mitotic figure. The mean and standard deviation of the number of tiles across 5 splits was 116,671 $\pm$ 2,389 for training, 3,354 $\pm$ 246 for validation, and 3,300 $\pm$ 196 for testing.

\subsection{Vision-language models}
\subsubsection{CLIP \cite{radford2021learning}}
Given a batch of $N$ pairs of (image, text), CLIP jointly trains an image encoder and text encoder to maximize the cosine similarity of embeddings between the matched $N$ pairs of (image, text) and minimize the cosine similarity of the unmatched $N^2 - N$ pairs. Specifically, CLIP is trained with a symmetric cross-entropy loss over the similarity scores derived by the InfoNCE loss:

\begin{equation}
\begin{split}
L = -\frac{1}{2N} \left(\sum_{i = 1}^{N} \log \frac{e ^{cos (\boldsymbol{u_i}, \boldsymbol{v_i)/\tau}}}{\sum_{j = 1}^{N} e ^{cos (\boldsymbol{u_i}, \boldsymbol{v_j)/\tau}}} \right. \\ \left. + \sum_{i = 1}^{N} \log \frac{e ^{cos (\boldsymbol{u_i}, \boldsymbol{v_i)/\tau}}}{\sum_{j = 1}^{N} e ^{cos (\boldsymbol{u_j}, \boldsymbol{v_i)/\tau}}}\right) \notag
\end{split}
\end{equation}
where $\tau$ is a temperature parameter to scale the logits,  $\boldsymbol{u_i}$ and $\boldsymbol{v_i}$ are embeddings produced by the image encoder and text encoder for the $ith$ image and text. 

The image encoder is a ViT-B/32 and the text encoder is a GPT-2 and neither has been pre-trained. CLIP was trained on more than 400 million image-text pairs from the Internet. During inference of image classification, the image is fed into the trained CLIP and the prediction label is the text from the most probable (image, text) pair. When using the MIDOG22 dataset for finetuning or zero-shot prediction, the text/label for mitotic tiles is ``mitotic" and the one for non-mitotic tiles is ``nonmitotic".

\subsubsection{BLIP \cite{li2022blip}} 
The pre-training of BLIP consists of an image encoder initialized from ViT-B/16 pre-trained on ImageNet data, two text encoders initialized from BERT-Base, and one text decoder initialized from BERT-Base. They are jointly trained with three losses that aim to promote vision-language alignment: image-text contrastive learning loss, image-text matching loss, and image-conditioned language modeling loss. The pre-training dataset consists of around 129 million image-text samples from six different datasets.

As for the downstream models, the image caption model consists of an image encoder and an image-grounded text decoder, and both are initialized from the BLIP pre-training step (Figure \ref{fig:figure1}c); the VQA model consists of an image encoder (Figure \ref{fig:figure1}b), image-grounded question encoder, and answer decoder, which are all initialized from the BLIP pre-training step. Both the image caption model and the VQA model learn to maximize the conditional likelihood of the output text $y$ under the forward autoregressive factorization:

\begin{equation}
L = - \sum_{i = 1}^{N}\sum_{t = 1}^{T_i} \log \, p(y_{i, t} | y_{i, 0:t-1}, x_i) \notag
\end{equation}
where $T_i$ is the number of word tokens in the $ith$ caption or answer and $x_i$ is the image embeddings for the image caption model and image-question embeddings for the VQA model. 

The BLIP image caption model was trained on 113k samples from the COCO dataset, and the BLIP VQA model was trained on 83k samples from the VQA2.0 dataset. For finetuning or zero-shot prediction tasks on the MIDOG22 data, two types of image captions were explored. BLIP binary image caption model's caption for mitotic tiles is ``mitotic" and the one for non-mitotic tiles is ``nonmitotic"; BLIP complete caption model (Figure \ref{fig:figure1}c)'s caption is ``[mitotic label], [tuomr type], [species], [scanner]". No prompt was used for binary and complete caption models. For BLIP VQA model, the question is ``This is an image of [species] [tumor type] taken using scanner [scanner]. Is there mitosis in the image?", and the answer is ``yes" for the mitotic tiles and ``no" for the non-mitotic tiles.

\subsection{Baseline models}
The effectiveness of the vision-language models was compared against several baseline models that incorporate different pre-training and finetuning strategies, as shown in Table \ref{table: table1}. Stain prediction is a pathology-specific self-supervised learning (SSL) task where a ResUNet was used to predict the H-stain from the E-stain \cite{koohbanani2021self}. SimSiam is a popular contrastive learning method that learns transformation invariant features from the unlabeled data \cite{chen2021exploring}. Both stain prediction and SimSiam models used a ResNet-50 \cite{he2016deep} as the backbone. The learned weights were transferred to the downstream ResNet-50 classification model for finetuning. A ResNet-50 with randomly initialized weights and with ImageNet-pre-trained weights were also included as baselines as was an ImageNet-pre-trained ViT-B/16 classifier \cite{dosovitskiy2020image}.

\subsection{Model training and evaluation}
All the baseline downstream models and the stain prediction pre-training model were trained with a batch size of 32, a learning rate of 0.0001, and the Adam optimizer. For SimSiam, the batch size was 128, and the learning rate was 0.005, and the optimizer was stochastic gradient descent with momentum of 0.9. All BLIP models were finetuned with a batch size of 32 and learning rate of 0.00001. CLIP was finetuned with a batch size of 512 and a learning rate of 0.0001. Both BLIP and CLIP models used the AdamW optimizer and cosine learning rate scheduler with warmup. These hyperparameters and optimizers were chosen by referencing the most commonly used ones in the literature and making minor adjustments around these common settings based on empirical experimentation.

The performance of all models was measured by the F1 score and area under the ROC curve (AUC) on the test sets across 5 different random splits (see Section \ref{sec:dataset}). 

\section{Results and discussion}

\begin{table*}[]
\caption{Mean F1 score and AUC on test sets for all baseline models and best vision-language model. SD: standard deviation. SSL: self-supervised learning. *: paired t-test p $<$ 0.05 compared to the other vision-only models}
\vspace{0.2cm}
\centering{

\begin{tabular}{@{}lllll@{}}
\toprule
Pre-training   & Finetuning & F1 score (SD) & AUC (SD)             \\ \midrule
None          & ResNet50 & 0.816 (0.0158) & 0.813 (0.0144)         \\
ResNet50 on ImageNet data & ResNet50 & 0.832 (0.0085) & 0.831 (0.0120) \\
SSL stain prediction on MIDOG22 & ResNet50 & 0.806 (0.0150) & 0.810 (0.0118) \\
SSL SimSiam on MIDOG22 & ResNet50 & 0.799 (0.0116) & 0.800 (0.0109) \\
ViT-B/16 on ImageNet data & ViT-B/16 & 0.821 (0.0127) & 0.813 (0.0121) \\
BLIP VQA on VQA2.0 data & BLIP VQA         & \textbf{0.860 (0.00941)*} & \textbf{0.853 (0.0116)*}  \\ \bottomrule
\end{tabular}

}
\label{table: table1}
\end{table*}

\begin{table*}[]
\caption{Mean F1 score and AUC on test sets for all vision-language models. SD: standard deviation.}
\vspace{0.2cm}
\centering{

\begin{tabular}{@{}lllll@{}}
\toprule
Pre-training   & Finetuning & F1 score (SD) & AUC (SD)             \\ \midrule
CLIP on Internet crawled data & None (zero-shot) & 0.675 (0.0127) & 0.499 (0.0123) \\
CLIP on Internet crawled data & CLIP & 0.798 (0.0118) & 0.782 (0.0128) \\
BLIP VQA on VQA2.0 data          & None (zero-shot) & 0.695 (0.00952) & 0.5 (0)         \\
BLIP VQA on VQA2.0 data & None (zero-shot) w/metadata & 0.704 (0.00898) & 0.529 (0.0201) \\
BLIP image caption on COCO data &BLIP binary caption & 0.855 (0.00898) & 0.838 (0.0142) \\
BLIP image caption on COCO data &BLIP complete caption & 0.852 (0.0113) & 0.845 (0.0133)\\
BLIP VQA on VQA2.0 data & BLIP VQA         & \textbf{0.860 (0.00941)} & \textbf{0.853 (0.0116)}  \\ \bottomrule
\end{tabular}

}
\label{table: table2}
\end{table*}

As shown in Table \ref{table: table1}, among the five baseline models, the ImageNet-pre-trained ResNet-50 had the best F1 score and AUC. The two SSL-pre-trained ResNet-50 models did not outperform the one trained with a random initialization. This indicates that large labeled datasets such as ImageNet might still be more desirable for pre-training. ViT-B/16 outperformed all other baseline models except for ImageNet-pre-trained ResNet50. 

Table \ref{table: table2} shows results for all vision-language models. The series of BLIP models outperformed the CLIP models, regardless of whether the model was finetuned. This makes sense since BLIP was pre-trained using three different objectives that all aim to promote image-text alignment while CLIP only had a contrastive-based objective. In addition, BLIP improved the pre-training data quality by leveraging synthetic captions and a caption filter that removes noisy captions, while CLIP only had the unfiltered captions that might be suboptimal for vision-language learning. Among BLIP models, adding metadata (tumor type, species, scanner) in the question improved the model performance. This was reflected in both the zero-shot setting and the finetuning setting where the BLIP VQA model had better F1 score and AUC as compared to BLIP binary caption and complete caption models which did not use metadata. The metadata served as prior knowledge or context, which might be informative of the prediction task and therefore improved the performance.

All three BLIP finetuned models (Table \ref{table: table2}) had statistically significantly higher F1 score and AUC as compared to the five baseline models (Table \ref{table: table1}), demonstrating the benefit of leveraging multimodal vision and language information. For the BLIP complete caption model, the results in Table \ref{table: table2} were calculated only on the predicted mitotic label for the purpose of comparison with other models. The average accuracy of the model predicting the entire caption correctly in the test sets was 0.740 $\pm$ 0.0222, which means BLIP is capable of predicting not only the label of interest but also the tile's associated metadata. CLIP did not outperform any of the vision-only baselines, indicating language does not provide additional value when not properly integrated with the visual information.

While providing promising performance, we note that the scope of this work is limited to binary tile classification. We leave more granular classification such as mitotic abnormality or mitotic phase to future work. A whole-slide level mitotic density will also be derived by running the trained models on overlapping tiles in a sliding window approach. In addition, more state-of-the-art large vision-language models and different prompting strategies will be explored.

\section{Conclusion}
In summary, we propose a new framework of mitosis detection using pre-trained large vision-language models by formulating the task as an image captioning task and a VQA task. We show that the BLIP image caption and VQA models outperform all vision-only baseline models, and adding metadata including tumor type, species, and scanner information can further improve performance. This pipeline can be extended to other prediction tasks, especially when highly informative clinical metadata is present. 
\section{Compliance with ethical standards}
This study was conducted using publicly available data. Additional approval was not required.  

\section{Conflict of Interests}
No external funding was received for conducting this study. The authors have no relevant financial or non-financial interests to disclose
 
\bibliographystyle{ieeetr}
\bibliography{refs}

\end{document}